\documentclass[acmsmall,screen]{acmart}

\AtBeginDocument{%
  }

\usepackage{graphicx}
\usepackage{amsmath}
\usepackage{amssymb}
\usepackage{amsthm}
\usepackage{booktabs}
\usepackage{algorithm}
\usepackage{algorithmic}
\usepackage{dsfont}
\usepackage{multirow}
\usepackage{bm}
\usepackage{pifont}
\usepackage{balance}
\usepackage{subfigure}
\usepackage{color}
\usepackage{xcolor}
\usepackage{colortbl}
\usepackage{adjustbox}

\newcommand\blfootnote[1]{%
  \begingroup
  \renewcommand\thefootnote{}\footnote{#1}%
  \addtocounter{footnote}{-1}%
  \endgroup
}

\setcopyright{acmlicensed}
\copyrightyear{2018}
\acmYear{2018}
\acmDOI{XXXXXXX.XXXXXXX}

\acmJournal{JACM}
\acmVolume{37}
\acmNumber{4}
\acmArticle{111}
\acmMonth{8}

\begin{document}

\title{Alleviating Hallucination in Large Vision-Language Models with Active Retrieval Augmentation}

\author{Xiaoye Qu\textsuperscript{$\dag$}}
\email{xiaoye@hust.edu.cn}
\affiliation{%
  \institution{Huazhong University of Science and Technology}
  \country{China}
}

\author{Qiyuan Chen\textsuperscript{$\dag$}}
\email{chenqiyuan1012@gmail.com}
\affiliation{%
  \institution{Zhejiang University}
  \country{China}
}

\author{Wei Wei\textsuperscript{$*$}}
\email{weiw@hust.edu.cn}
\affiliation{%
  \institution{Huazhong University of Science and Technology}
  \country{China}
}

\author{Jiashuo Sun}
\email{gasolsun36@gmail.com}
\affiliation{%
  \institution{Xiamen University}
  \country{China}
}

\author{Jianfeng Dong}
\email{dongjf24@gmail.com}
\affiliation{%
  \institution{Zhejiang Gongshang University}
  \country{China}
}

\renewcommand{\shortauthors}{Qu et al.}

\begin{abstract}
Despite the remarkable ability of large vision-language models (LVLMs) in image comprehension, these models frequently generate plausible yet factually incorrect responses, a phenomenon known as hallucination. 
Recently, in large language models (LLMs), augmenting LLMs by retrieving information from external knowledge resources has been proven as a promising solution to mitigate hallucinations. 
However, the retrieval augmentation in LVLM significantly lags behind the widespread applications of LVLM. 
Moreover, when transferred to augmenting LVLMs, sometimes the hallucination degree of the model is even exacerbated. 
Motivated by the research gap and counter-intuitive phenomenon, 
we introduce a novel framework, the Active Retrieval-Augmented large vision-language model (ARA), specifically designed to address hallucinations by incorporating three critical dimensions: 
(i) dissecting the retrieval targets based on the inherent hierarchical structures of images. (ii) pinpointing the most effective retrieval methods and filtering out the reliable retrieval results. (iii) timing the retrieval process to coincide with episodes of low certainty, while circumventing unnecessary retrieval during periods of high certainty. To assess the capability of our proposed ARA model in reducing hallucination, we employ three widely used LVLM models (LLaVA-1.5, Qwen-VL, and mPLUG-Owl2) across four benchmarks. 
Our empirical observations suggest that by utilizing fitting retrieval mechanisms and timing the retrieval judiciously, we can effectively mitigate the hallucination problem. 
We hope that this study can provide deeper insights into how to adapt the retrieval augmentation to LVLMs for reducing hallucinations with more effective retrieval and minimal retrieval occurrences.
\end{abstract}

\blfootnote{
\textsuperscript{$\dag$}The first two authors contribute equally.\\
\textsuperscript{$*$}Corresponding author.
}

%   <concept_desc>Information systems~Multimedia and multimodal retrieval</concept_desc>
%   <concept_desc>Information systems~Video search</concept_desc>

% \ccsdesc[500]{Computing methodologies~Artificial intelligence}
% % \ccsdesc[300]{Computing methodologies~Video search}

% %%
% %% Keywords. The author(s) should pick words that accurately describe
% %% the work being presented. Separate the keywords with commas.
% \keywords{Large Vision-Language Models, Hallucination, Retrieval Augmentation}


\maketitle

\section{Introduction}

Recently, Large Vision-Language Models (LVLMs) ~\cite{bai2023qwen,liu2024visual,ye2023mplug,dai2024instructblip,zhu2023minigpt,chen2023minigpt,liu2024survey} have gained widespread application across various scenarios due to their significant capacity to generate contextually appropriate text for visual inputs. 
Benefiting from the advancements in model architecture, training methods, and data diversity, LVLMs exhibit superior performance for various tasks such as visual question answering~\cite{guo2022images,prasad2023rephrase} and image captioning~\cite{ramos2023smallcap,zeng2024meacap}.
Despite these advancements, LVLMs still suffer from a significant challenge termed “hallucination”, whereby the models produce semantically plausible but factually inaccurate text, misaligned with the ground-truth content of the associated image ~\cite{zhou2023analyzing,yin2023woodpecker,leng2023mitigating,huang2023opera,chen2024halc,zhu2024ibd}.
This problem damages the practical employment of LVLMs, particularly in fields that require accurate content generation, such as medical~\cite{wang2023chatcad,li2024llava} and robotics~\cite{shah2023lm,liu2024online}, potentially resulting in severe consequences. 
Hence, addressing the hallucination problem is paramount for bolstering the credibility of LVLMs across real-world scenarios.

\begin{figure*}
\centering
\includegraphics[width=0.95\linewidth]{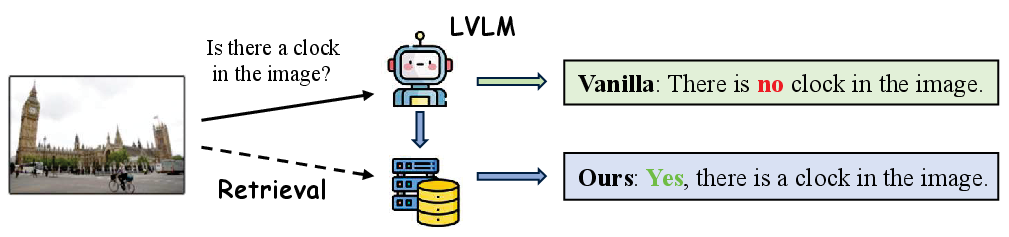}
\caption{Given an input image and corresponding query, the vanilla LVLM fails to identify the clock within the image and outputs a hallucinated answer. 
However, by employing the retrieval augmentation strategy proposed in this paper, the LVLM model can accurately answer the query.
}
\label{fig1}
\end{figure*}

To mitigate hallucinations in LVLMs, a series of attempts have been recently proposed and can be broadly classified into two categories. 
The first class of approach retrains the LVLMs with constructed hallucination-related datasets by
supervised fine-tuning (SFT) ~\cite{liu2023aligning,gunjal2024detecting,wang2024mitigating}, or Reinforcement Learning from Human Feedback (RLHF)~\cite{sun2023aligning,yu2023rlhf}.
Although effective, these methods introduce significantly additional training costs. 
The other solutions focus on designing more robust decoding strategies~\cite{chen2024halc,huang2023opera,yin2023woodpecker,leng2023mitigating}. For example, VCD ~\cite{leng2023mitigating} introduces visual contrastive decoding to contrast output distributions derived from original and distorted visual inputs. 
In this way, the LVLM reduces the over-reliance on statistical bias and unimodal prior.
Although these methods are training-free, they still suffer from the limitations of LVLMs’ static parametric capacity. 
Recently, in the realm of large language models (LMs), augmenting LLMs \cite{karpukhin2020dense,izacard2022atlas,ram2023context} by retrieving information from external knowledge resources has shown promise in reducing language hallucinations. Furthermore, this retrieval augmentation serves as a flexible way to extend beyond the model's inherent knowledge without the burden of extensive training costs.
In this paper, we aim to propose a novel framework to augment LVLMs with external knowledge by introducing an Active Retrieval-Augmented large vision-language model (ARA) for mitigating hallucinations. As shown in Figure \ref{fig1}, given an input image and query, the LVLM tends to produce hallucinated answers. Instead, our model can accurately identify the most relevant pairs and output the correct answer.

Concretely, our proposed ARA is grounded in three critical dimensions for mitigating hallucinations in LVLMs. 
Firstly, given the inherently hierarchical nature of images, simple full-image retrieval may result in noise and irrelevant outcomes, thus it is imperative to decompose the target object causing hallucination from the input image for more accurate retrieval. 
Secondly, in contrast to augmenting LLMs, LVLMs offer diverse retrieval modalities due to their multimodal inputs. 
In this way, discerning the most effective retrieval technique and securing trustworthy results is crucial for ensuring retrieving performance. 
Lastly, excessive retrieval activations may lead to undue time expenditure and unnecessary retrieval.
By initiating retrieval only during periods of low LVLM certainty and knowledge deficiency, one can circumvent unnecessary or improper information retrieval.

Based on the above analysis, given the input image and query, our proposed ARA first devises a coarse-to-fine retrieving paradigm. 
The target object is first extracted from the input query and subsequently positioned within a particular region of the input image. To build a robust retrieving system, we simultaneously engage in coarse-grained and fine-grained retrieval process which retrieves the full image as well as the specific regions pertaining to the target object. Recognizing the plurality of retrieval strategies due to the multimodal input, an extensive analysis of different methods is conducted to determine the most optimal retrieving approach. 
Subsequent to acquiring pertinent text and image pairs from external databases, a reranking strategy is employed to eliminate unreliable outcomes, thereby augmenting LVLMs with the appropriate external information.
Particularly, retrieval operations are circumvented when the LVLM exhibits high certainty, to prevent the inclusion of superfluous retrieval that may introduce redundant information. 
In this context, retrieval is activated using a difficulty metric that relies on the mutual information between multimodal inputs.

To assess the efficacy of ARA, empirical evaluations are performed using three prevalent LVLMs across four benchmarks related to hallucination challenges.
The promising results demonstrate that our framework can effectively mitigate hallucinations. 
To sum up, our contributions are summarized as follows:

\begin{itemize}
    \item Our research thoroughly investigates the three critical dimensions of augmenting LVLMs with external knowledge, including accurately dissecting the retrieval targets, pinpointing the most effective retrieval methods, and the sensible triggering of the retrieval process. 
    
    \item We introduce an Active Retrieval-Augmented large vision-language model (ARA) for mitigating hallucination. 
    Our empirical evidence indicates that, by employing appropriate retrieval mechanisms and carefully timing the retrievals, we can significantly address the issue of hallucination while maintaining a moderate frequency of retrieval.

    \item Through comprehensive experiments, we demonstrate the superior performance of our proposed ARA in alleviating hallucinations using three well-known LVLMs across four hallucination-related benchmarks. The results in our paper also suggest that with optimal retrieval settings, the potential of augmenting LVLMs can be more effectively harnessed.
\end{itemize}

\section{Related Work}

\subsection{Large Visual-Language Models}

Recently, large-scale models have attracted significant attention and gained widespread applications \cite{wang2024twin,zhuounified,lu2024mitigating,su2024timo,su2024living}. 
LVLMs significantly enhance the interaction between humans and AI in a more natural and intuitive
ways and demonstrate remarkable capabilities in understanding and generating multi-modal content.
With the aid of advanced Large Language Models like LLaMA~\cite{touvron2023llama} and Qwen~\cite{bai2023qwenlm}, a batch of LVLMs such as LLaVA-1.5 \cite{liu2024visual}, Qwen-VL \cite{bai2023qwen}, and mPLUG-Owl2 \cite{ye2023mplug} have emerged, which can comprehend and generate a wide array of content by utilizing information from distinct modalities like texts and images. 
The training of LVLMs is divided into two critical phases: feature alignment and instruction fine-tuning. The former aims to model visual and textual inputs coherently, while the latter strives to enhance the ability of LVLMs to comprehend user queries. 
Despite the advancements, these LVLMs continue to face significant challenges with hallucination issues. Thus, in this paper, we focus on solving hallucination problems to promote the use of LVLMs in practical scenarios.

\subsection{Hallucination in LVLMs}

In LVLMs, hallucination refers to models that generate seemingly plausible outputs inclusive of objects,
attributes, or relations that do not correspond with the images \cite{yin2023survey,li2023evaluating,zhai2023halle}.
Approaches to address hallucinations in LVLMs have largely fallen into two camps. The first emphasizes model refinement through supervised fine-tuning or application of reinforcement learning techniques. For instance, LRV \cite{liu2023mitigating} seeks to diminish hallucinatory outputs by utilizing broad and varied supervised fine-tuning datasets. 
For methods based on reinforcement learning, LLaVA-RLHF \cite{sun2023aligning} is the pioneer in applying Reinforcement Learning with Human Feedback (RLHF) to mitigate hallucination in LVLMs. 
This approach is further refined by RLHF-V \cite{yu2023rlhf}, which incorporates detailed corrective human feedback.
Considering the instability and training difficulty of RLHF, Zhao et al. \cite{zhao2023beyond} employ Direct Preference Optimization (DPO) and build a hallucination-aware dataset for alleviating hallucination. 
Although 
While these methods have yielded notable advancements, they also introduce significant training overheads and are prone to overfitting the training data. 
The second category consists of training-free strategies designed to circumvent hallucination without incurring additional training expenses.
VCD \cite{leng2023mitigating} endeavors to rectify the model's over-reliance on linguistic priors and statistical leanings by comparing the output distributions from both unaltered and visually perturbed inputs.
OPERA \cite{huang2023opera} mitigates overconfidence in model predictions by integrating a penalty term into the model logits during beam search decoding, further enhanced with a strategic rollback feature. 
Although these methods are effective, they still suffer from the limitations of LVLMs’ static parametric capacity. 
In this paper, different from the above two paradigms, we focus on augmenting LVLMs with external knowledge to mitigate hallucination in LVLMs.  

\subsection{Retrieval-Augmented Generation}

In the realm of large language models (LLMs), Retrieval-Augmented Generation (RAG) has been widely used and shown promising results in mitigating hallucinations. RAG in LLMs retrieves relevant information from an external knowledge base before LLMs respond to a query~\cite{karpukhin2020dense,izacard2022atlas,ram2023context}, thereby enabling them to collaboratively generate responses by leveraging the retrieved external non-parameterized knowledge alongside their internal parameterized knowledge. 
However, its application in LVLMs has not been thoroughly explored. Since LVLMs present a multimodal nature, findings pertinent to LLMs cannot be indiscriminately extrapolated to LVLMs. 
Predominantly, existing research on retrieval augmentation in multimodal tasks has been confined to image captioning \cite{ramos2023smallcap,ramos2023lmcap,chen2023retrieval,sarto2024towards} or image generation \cite{chen2022re}, with a narrow focus that overlooks the broader implications of RAG in LVLMs, especially concerning its capacity for hallucination reduction.
In this paper, our research conducts an extensive evaluation of RAG's applicability in LVLMs to mitigate hallucination by tailoring retrieval strategies and optimizing active retrieval processes. 
It is worth noting that the active retrieving strategy has also been discussed in \cite{jiang2023active} in LLM. However, they propose a token-level confidence for triggering the retrieval, which is not applicable in LVLMs and we will investigate it in the ablation study.

\section{Method}

\begin{figure}[t!]
\centering
\includegraphics[width=0.95\linewidth]{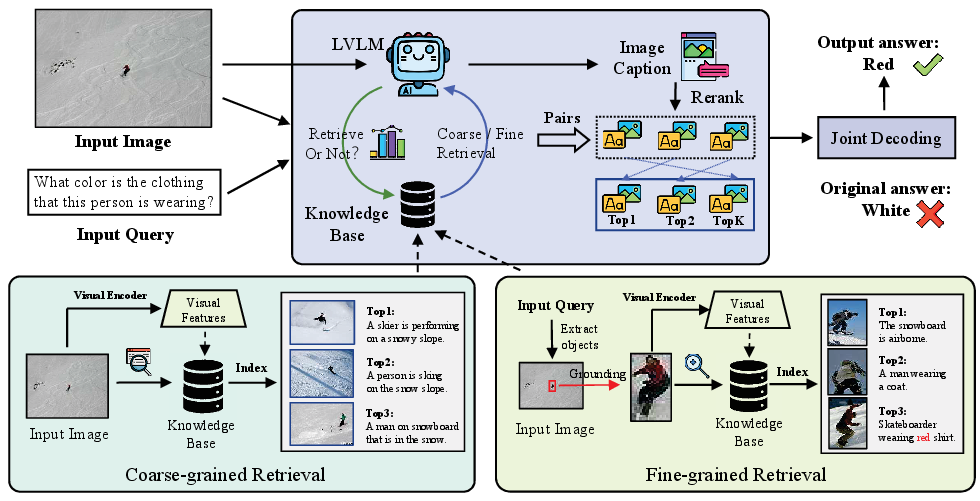}
\caption{The overall pipeline of our Active Retrieval-Augmented large vision-language model (ARA). Given the input image and query, our model first actively determines whether the input pair requires retrieval. If retrieval is not necessary, the LVLM will directly answer the input query based on the image. Otherwise, our model will perform a coarse-to-fine retrieval based on the input image. Subsequently, the retrieved image and text pairs are further reranked according to the caption of the input image. Finally, the retrieved coarse-grained and fine-grained pairs are jointly decoded to produce the final answer.}
\label{fig2}
\end{figure}

In this section, we first describe the generation procedure of the LVLMs, laying the groundwork for comprehending our approach. Subsequently, we will introduce our Active Retrieval-Augmentation large vision-language model (ARA) in detail. As shown in Figure \ref{fig2}, given the input image and query, the LVLM first analyzes them based on mutual information and determines whether to trigger the retrieval process. If the retrieval is necessary, our ARA will follow a coarse-to-fine paradigm to dissect the retrieval targets and conduct jointly coarse-grained and fine-grained retrieval. Following this, a reranking of the retrieval outcomes is performed for additional refinement. Finally, the LVLM harnesses this externally sourced knowledge to generate the final response.

\subsection{LVLM Input and Decoding}
 
The input of LVLMs contains both image and text. Typically, the LVLMs use a vision encoder (e.g. CLIP) to extract visual tokens from the input images, and then map them into the input space of LLMs with a linear projection module. 
Subsequently, the mapped visual tokens are used as part of the LLM input, along with the text input. 
In this paper, we denote the visual tokens as $\mathbf{x}^v = \{x_0,x_1,\ldots,x_{N-1}\}$. Here $N$ is the length of the visual tokens and it is a fixed number in most cases.  
Correspondingly, the input text is tokenized and we denote it as $\mathbf{x}^p = \{x_0,x_{1},\ldots,x_{M-1}\}$.
The image and text tokens are concatenated as the final input sequence and we denote it as  $\{x_i\}_{t=0}^{T-1}$ that $T=N+M$.

During decoding, LVLM gets the probabilities for the next token prediction, formally:
\begin{equation}
    p(x_t|x_{<t}) = \text{SoftMax}[{LVLM}(h_t)]_{x_t}, \quad x_t\in \mathcal{V},
\end{equation}
where $h_t$ denotes the output hidden states at time step t for the input image and query, $x_{<t}$ to simplify the sequence $\{x_i\}_{i=0}^{t-1}$ and $\mathcal{V}$ means the whole vocabulary set. For greedy decoding, the token with the highest probability will be chosen.

\subsection{Active Triggering Retrieval}

As shown in Figure \ref{fig2}, given the input image and query, our ARA model first determines whether the input pair requires retrieval. In this way, we can circumvent unnecessary retrieval during periods of high certainty. 
In this section, we propose three active retrieval methods to explore the difficulty metrics that influence when to retrieve, including Confidence-aware Active Retrieval, 
Question-aware Active Retrieval, and Image-aware Active Retrieval. The first method relies on the confidence of the model output, and the latter two matrices depend on the mutual information between the inputs.
An ideal metric should effectively reduce the frequency of retrievals and can easily apply to different LVLMs.

\noindent\textbf{Confidence-aware Active Retrieval}. We feed the image and query into LVLMs to generate the answer. If the LVLM is confident about the answer, we accept it without retrieving additional information. Specifically, we actively trigger retrieval if any token of the answer has a confidence lower than the threshold $\theta$. In this method, $\theta=0$ indicates retrieval will never be used and $\theta=1$ means that all inputs will use the retrieval. This strategy is inspired by \cite{jiang2023active} in LLM.

\noindent\textbf{Query-aware Active Retrieval}. To better evaluate the difficulty of queries from a view of language bias, we evaluate the degree of LVLMs answering relying on the input image. To this end, we define a difficulty metric using mutual information based on the difference below:

\begin{equation}
M_{ij} = \log \frac{P(a_{ij} | V_i, Q_i)}{P(a_{ij})} - \log \frac{P(a_{ij} | Q_i)}{P(a_{ij})} = \log \frac{P(a_{ij} | V_i, Q_i)}{P(a_{ij} | Q_i)}. 
\end{equation}

where $P(a_{ij} | Q_i)$ indicates the output probability with only query input to the LVLMs and $P(a_{ij} | V_i, Q_i)$ means the output probability with both query and image as model input. The higher $M_{ij}$ indicates more visual reasoning is needed to answer. $M_{ij}<0$ illustrates that the answer excessively depends on the input query. Thus, we define a threshold $\theta$ and trigger retrieval if $M_{ij}<\theta$, where insufficient visual information is provided for answering the query. Incorporating retrieved information will enhance the confidence of LVLMs to answer.  

\noindent\textbf{Image-aware Active Retrieval}. Similar to the above query-aware active retrieval, we use the probability difference between the original and noisy image to evaluate the image information to answer the query. During the implementation, we use the method in \cite{leng2023mitigating} to add noise.

\begin{equation}
D_{ij} = \log \frac{P(a_{ij} | V_i, Q_i)}{P(a_{ij})} - \log \frac{P(a_{ij} | V'_i, Q_i)}{P(a_{ij})} = \log \frac{P(a_{ij} | V_i, Q_i)}{P(a_{ij} | V'_i, Q_i)}. 
\end{equation}

where $P(a_{ij} | V_i, Q_i)$ indicates the output probability with query and input image, $P(a_{ij} | V'_i, Q_i)$ means the output probability with query and noisy image. Analogously, we define a boundary $\theta$ and trigger retrieval if $D_{ij}<\theta$.

\subsection{Coarse-to-Fine Hierarchical Retrieval}

Given the inherently hierarchical nature of images, simple full-image retrieval may result in noise and irrelevant outcomes. As shown in Figure \ref{fig2}, using the full image for retrieval can not obtain the most relevant information for reasoning, namely ``red shirt". Thus it is imperative to decompose the target object causing hallucination from the input image for more accurate retrieval. 
In the following, we will first describe the coarse-grained retrieval and then introduce the fine-grained retrieve. 

\noindent\textbf{Coarse-grained Retrieve.} 
Initially, based on the input image, we use the CLIP to extract the visual embedding. Similarly, we build the embeddings for the images and corresponding texts in the knowledge base. Then, we can retrieve the relevant images from knowledge bases with the similarity between images. Finally, the top-K image and text pairs will be fetched for the following reasoning. 
Benefiting from the multimodal encoding ability of CLIP, we can also retrieve the top-K pairs by matching the input image with texts in the database. We will compare different retrieving methods in the ablation study.
% In this stage, we can obtain a paired caption with the retrieved image. 

\noindent\textbf{Fine-grained Retrieve.} While full-image retrieval can incorporate valuable information, it may overlook fine-grained details, such as diminutive objects and intricate attribute data. Therefore, it is crucial to propose a fine-grained retrieval mechanism that concentrates on the distinct objects within an image. 
To conduct fine-grained retrieval, we initially extract key entities from the input query utilizing the large language model LLaMA2-7B \cite{touvron2023llama}. This extraction is facilitated by in-context learning \cite{dong2022survey}, where demonstrations are presented to the LLM, which subsequently outputs the specific entities from the input query. 
Following this, Grounding Dino \cite{liu2023grounding} is deployed to identify the objects in the image that correspond to the extracted entities. As illustrated in Figure \ref{fig2}, we crop the targeted object and then perform retrieval in a manner akin to the previously described coarse-grained retrieval.

\subsection{Reranking Retrieving Results}

Throughout the above retrieving process, the similarity is computed by the CLIP embeddings between images. In this way, the visually similar but semantically different pairs may be recalled. 
To avoid noisy retrieving results which bring different semantic, after obtaining the retrieved pairs for both coarse-grained and fine-grained retrieval, we subsequently propose a reranking strategy designed to recalibrate the retrieved outcomes. 
Specifically, we adopt the LVLM to describe the input image and generate a detailed caption. For coarse-grained retrieval, the input full image is characterized, while describing the cropped image for the fine-grained retrieval.
In this way, we are able to derive the semantics of the input image.
% Simultaneously, we can obtain the semantics for the retrieved images. 
Then, we extract the textual embeddings for both image captions and captions of retrieved images with the CLIP. Finally, we can rerank the retrieved pairs according to the semantic similarity between these captions. In this stage, we also explore other reranking methods, such as k-reciprocal encoding \cite{zhong2017re} or using an external visual entailment tool \cite{Liu2022VisualSR}. We will compare them in the ablation study.

\subsection{Joint Coarse-grained and Fine-grained Decoding}

After retrieving and ranking the pairs, we can conduct a joint coarse-grained and fine-grained decoding to yield the final responses. According to the different ways of using the retrieved results, we have two schemes including probability-level fusion and instance-level fusion. The former integrates the results of coarse-grained and fine-grained retrieval into a single prompt for inference, while the latter conducts separate inferences on the coarse-grained and fine-grained retrieval results, followed by a probabilistic-level weighting.

\noindent\textbf{Probability-level Fusion.} In this paradigm, we use the prompt for both coarse-grained retrieval and fine-grained retrieval: ``Here are the image-caption pairs similar to the test image: $\langle$Retrieval Pairs$\rangle$. Based on these pairs and this image: $\langle$Input Image$\rangle$. Answer this question: $\langle$Input query$\rangle$". 
Here we denote the retrieval-augmented LVLMs as below:

\begin{equation}
    p^{coarse}(x_t|x_{<t}) = \text{SoftMax}[{LVLM}([R_I:h_t])]_{x_t}, \quad x_t\in \mathcal{V},
\end{equation}

where $R_I$ indicates the hidden states of retrieved image-level pairs. For simplicity, we directly use [:] here. We use $p^{coarse}$ to represent decoding with coarse-grained retrieval. Similarly, 

\begin{equation}
    p^{fine}(x_t|x_{<t}) = \text{SoftMax}[{LVLM}([R_O:h_t])]_{x_t}, \quad x_t\in \mathcal{V},
\end{equation}

% Similar to the above coarse-grained retrieval, 
where $R_O$ indicates the retrieved fine-grained object-level pairs. We represent the decoding with object-level fine retrieval by $p^{fine}(x_t|x_{<t})$. For the probability-level fusion, we can combine them with a hyperparameter $\alpha$ in [0,1] to control the importance of two kinds of retrieval. 

\begin{equation}
    p^{coarse\&fine}(x_t|x_{<t}) = \alpha p^{coarse}(x_t|x_{<t}) + (1-\alpha)p^{fine}(x_t|x_{<t}),
\end{equation}

\noindent\textbf{Instance-level Fusion.} 
For this kind of fusion, we can integrate both coarse-grained retrieval and fine-grained retrieval with prompts: ``Here are the image-caption pairs similar to the test image: $\langle$Coarse-grained Retrieval Pairs$\rangle$. Here are the image-caption pairs: $\langle$Fine-grained Retrieval Pairs$\rangle$ similar to the $\langle$entity$\rangle$ in the input image. Based on these pairs and this input image: $\langle$Input Image$\rangle$. Answer this question: $\langle$Input query$\rangle$." Here we extract the entity from the input query and fill it in the prompt. 
 
\begin{equation}
    p^{coarse+fine}(x_t|x_{<t}) = \text{SoftMax}[{LVLM}([R_I:R_O:h_t])]_{x_t}, \quad x_t\in \mathcal{V},
\end{equation}

\begin{table*}[tp]
\centering
\caption{Results on POPE. \textit{Regular} decoding denotes greedy decoding, whereas \textit{RAR} refers to our method with augmentation.  
% The best performances within each setting are \textbf{bolded}.
}
\resizebox{0.85\linewidth}{!}{%
\begin{tabular}{cllllll|l}
\hline
\textbf{Dataset}          & \textbf{Setting}                         & \textbf{Model}                & \textbf{Decoding} & Accuracy$\uparrow$ & Precision & Recall & F1 Score$\uparrow$  \\ \hline
\multirow{18}{*}{MSCOCO}  & \multirow{6}{*}{\textit{Random}}      & \multirow{2}{*}{LLaVA1.5}     & Regular           & 86.50 & 97.40 & 75.00 & 84.75                                          \\
                          &                                       &                               & \cellcolor{gray!20}RAR               & \cellcolor{gray!20}89.43 & \cellcolor{gray!20}95.96 & \cellcolor{gray!20}82.33 & \cellcolor{gray!20}88.63 \\
                          &                                       & \multirow{2}{*}{Qwen-VL}     & Regular           & 82.13 & 98.88 & 65.00 & 78.44  \\
                          &                                       &                               & \cellcolor{gray!20}RAR               & \cellcolor{gray!20}84.27 & \cellcolor{gray!20}98.22 & \cellcolor{gray!20}69.80 & \cellcolor{gray!20}81.61  \\
                          &                                       & \multirow{2}{*}{mPLUG-Owl2} & Regular           & 85.97 & 95.15 & 75.80 & 84.38  \\
                          &                                       &                               & \cellcolor{gray!20}RAR               & \cellcolor{gray!20}89.60 & \cellcolor{gray!20}94.33 & \cellcolor{gray!20}84.27 & \cellcolor{gray!20}89.01  \\ \cline{2-8} 
                          & \multirow{6}{*}{\textit{Popular}}     & \multirow{2}{*}{LLaVA1.5}     & Regular           & 85.53 & 95.02 & 75.00 & 83.83  \\
                          &                                       &                               & \cellcolor{gray!20}RAR               & \cellcolor{gray!20}87.47 & \cellcolor{gray!20}92.07 & \cellcolor{gray!20}82.00 & \cellcolor{gray!20}86.74  \\
                          &                                       & \multirow{2}{*}{Qwen-VL}     & Regular           & 81.93 & 98.29 & 65.00 & 78.25  \\
                          &                                       &                               & \cellcolor{gray!20}RAR               & \cellcolor{gray!20}84.03 & \cellcolor{gray!20}97.58 & \cellcolor{gray!20}69.80 & \cellcolor{gray!20}81.38  \\
                          &                                       & \multirow{2}{*}{mPLUG-Owl2} & Regular           & 84.63 & 92.06 & 75.80 & 83.14  \\
                          &                                       &                               & \cellcolor{gray!20}RAR               & \cellcolor{gray!20}86.10 & \cellcolor{gray!20}87.68 & \cellcolor{gray!20}84.00 & \cellcolor{gray!20}85.80  \\ \cline{2-8} 
                          & \multirow{6}{*}{\textit{Adversarial}} & \multirow{2}{*}{LLaVA1.5}     & Regular           & 83.60 & 90.71 & 74.87 & 82.03  \\
                          &                                       &                               & \cellcolor{gray!20}RAR               & \cellcolor{gray!20}84.53 & \cellcolor{gray!20}87.37 & \cellcolor{gray!20}80.73 & \cellcolor{gray!20}83.92  \\
                          &                                       & \multirow{2}{*}{Qwen-VL}     & Regular           & 80.97 & 95.41 & 65.07 & 77.37  \\
                          &                                       &                               & \cellcolor{gray!20}RAR               & \cellcolor{gray!20}82.90 & \cellcolor{gray!20}94.58 & \cellcolor{gray!20}69.80 & \cellcolor{gray!20}80.32  \\
                          &                                       & \multirow{2}{*}{mPLUG-Owl2} & Regular           & 82.37 & 87.32 & 75.73 & 81.11  \\
                          &                                       &                               & \cellcolor{gray!20}RAR               & \cellcolor{gray!20}82.77 & \cellcolor{gray!20}82.40 & \cellcolor{gray!20}83.33 & \cellcolor{gray!20}82.86  \\ \hline
\multirow{18}{*}{A-OKVQA} & \multirow{6}{*}{\textit{Random}}      & \multirow{2}{*}{LLaVA1.5}     & Regular           & 88.63 & 93.54 & 83.00 & 87.95  \\
                          &                                       &                               & \cellcolor{gray!20}RAR               & \cellcolor{gray!20}90.27 & \cellcolor{gray!20}90.81 & \cellcolor{gray!20}89.60 & \cellcolor{gray!20}90.20  \\
                          &                                       & \multirow{2}{*}{Qwen-VL}     & Regular           & 83.60 & 95.90 & 70.20 & 81.06  \\
                          &                                       &                               & \cellcolor{gray!20}RAR               & \cellcolor{gray!20}86.53 & \cellcolor{gray!20}95.74 & \cellcolor{gray!20}76.47 & \cellcolor{gray!20}85.03  \\
                          &                                       & \multirow{2}{*}{mPLUG-Owl2} & Regular           & 86.47 & 91.07 & 80.87 & 85.66  \\
                          &                                       &                               & \cellcolor{gray!20}RAR               & \cellcolor{gray!20}88.60 & \cellcolor{gray!20}89.33 & \cellcolor{gray!20}87.67 & \cellcolor{gray!20}88.49  \\ \cline{2-8} 
                          & \multirow{6}{*}{\textit{Popular}}     & \multirow{2}{*}{LLaVA1.5}     & Regular           & 85.53 & 87.43 & 83.00 & 85.16  \\
                          &                                       &                               & \cellcolor{gray!20}RAR               & \cellcolor{gray!20}85.63 & \cellcolor{gray!20}85.85 & \cellcolor{gray!20}85.33 & \cellcolor{gray!20}85.59  \\
                          &                                       & \multirow{2}{*}{Qwen-VL}     & Regular           & 83.47 & 95.55 & 70.20 & 80.94  \\
                          &                                       &                               & \cellcolor{gray!20}RAR               & \cellcolor{gray!20}85.97 & \cellcolor{gray!20}94.40 & \cellcolor{gray!20}76.47 & \cellcolor{gray!20}84.49 \\
                          &                                       & \multirow{2}{*}{mPLUG-Owl2} & Regular           & 82.43 & 83.48 & 80.87 & 82.15   \\
                          &                                       &                               & \cellcolor{gray!20}RAR               & \cellcolor{gray!20}82.77 & \cellcolor{gray!20}80.17 & \cellcolor{gray!20}87.07 & \cellcolor{gray!20}83.48  \\ \cline{2-8} 
                          & \multirow{6}{*}{\textit{Adversarial}} & \multirow{2}{*}{LLaVA1.5}     & Regular           & 79.13 & 77.04 & 83.00 & 79.91  \\
                          &                                       &                               & \cellcolor{gray!20}RAR               & \cellcolor{gray!20}79.33 & \cellcolor{gray!20}76.96 & \cellcolor{gray!20}83.73 & \cellcolor{gray!20}80.20  \\
                          &                                       & \multirow{2}{*}{Qwen-VL}     & Regular           & 78.97 & 85.13 & 70.20 & 76.95  \\
                          &                                       &                               & \cellcolor{gray!20}RAR               & \cellcolor{gray!20}80.50 & \cellcolor{gray!20}83.18 & \cellcolor{gray!20}76.47 & \cellcolor{gray!20}79.68  \\
                          &                                       & \multirow{2}{*}{mPLUG-Owl2} & Regular           & 74.70 & 73.38 & 80.87 & 76.94  \\
                          &                                       &                               & \cellcolor{gray!20}RAR               & \cellcolor{gray!20}75.63 & \cellcolor{gray!20}71.54 & \cellcolor{gray!20}85.13 & \cellcolor{gray!20}77.75  \\ \hline
\multirow{18}{*}{GQA}     & \multirow{6}{*}{\textit{Random}}      & \multirow{2}{*}{LLaVA1.5}     & Regular           & 88.87 & 93.64 & 83.40 & 88.22 \\
                          &                                       &                               & \cellcolor{gray!20}RAR               & \cellcolor{gray!20}90.10 & \cellcolor{gray!20}91.57 & \cellcolor{gray!20}88.33 & \cellcolor{gray!20}89.92  \\
                          &                                       & \multirow{2}{*}{Qwen-VL}     & Regular           & 82.63 & 95.45 & 68.53 & 79.78  \\
                          &                                       &                               & \cellcolor{gray!20}RAR               & \cellcolor{gray!20}85.60 & \cellcolor{gray!20}96.27 & \cellcolor{gray!20}74.07 & \cellcolor{gray!20}83.72  \\
                          &                                       & \multirow{2}{*}{mPLUG-Owl2} & Regular           & 85.17 & 91.05 & 78.00 & 84.02  \\
                          &                                       &                               & \cellcolor{gray!20}RAR               & \cellcolor{gray!20}86.90 & \cellcolor{gray!20}89.51 & \cellcolor{gray!20}83.60 & \cellcolor{gray!20}86.45  \\ \cline{2-8} 
                          & \multirow{6}{*}{\textit{Popular}}     & \multirow{2}{*}{LLaVA1.5}     & Regular           & 84.57 & 85.39 & 83.40 & 84.38  \\
                          &                                       &                               & \cellcolor{gray!20}RAR               & \cellcolor{gray!20}84.70 & \cellcolor{gray!20}84.27 & \cellcolor{gray!20}85.33 & \cellcolor{gray!20}84.80  \\
                          &                                       & \multirow{2}{*}{Qwen-VL}     & Regular           & 80.40 & 89.86 & 68.53 & 77.76  \\
                          &                                       &                               & \cellcolor{gray!20}RAR               & \cellcolor{gray!20}83.43 & \cellcolor{gray!20}91.41 & \cellcolor{gray!20}73.80 & \cellcolor{gray!20}81.67  \\
                          &                                       & \multirow{2}{*}{mPLUG-Owl2} & Regular           & 78.67 & 79.05 & 78.0 & 78.52  \\
                          &                                       &                               & \cellcolor{gray!20}RAR               & \cellcolor{gray!20}79.70 & \cellcolor{gray!20}78.07 & \cellcolor{gray!20}82.60 & \cellcolor{gray!20}80.27  \\ \cline{2-8} 
                          & \multirow{6}{*}{\textit{Adversarial}} & \multirow{2}{*}{LLaVA1.5}     & Regular           & 81.63 & 80.55 & 83.40 & 81.95 \\
                          &                                       &                               & \cellcolor{gray!20}RAR               & \cellcolor{gray!20}81.83 & \cellcolor{gray!20}80.43 & \cellcolor{gray!20}84.13 & \cellcolor{gray!20}82.24  \\
                          &                                       & \multirow{2}{*}{Qwen-VL}     & Regular           & 78.40 & 85.38 & 68.53 & 76.04  \\
                          &                                       &                               & \cellcolor{gray!20}RAR               & \cellcolor{gray!20}80.93 & \cellcolor{gray!20}85.86 & \cellcolor{gray!20}74.07 & \cellcolor{gray!20}79.53  \\
                          &                                       & \multirow{2}{*}{mPLUG-Owl2} & Regular           & 76.40 & 75.58 & 78.00 & 76.77  \\
                          &                                       &                               & \cellcolor{gray!20}RAR               & \cellcolor{gray!20}76.53 & \cellcolor{gray!20}75.38 & \cellcolor{gray!20}78.80 & \cellcolor{gray!20}77.05  \\ \hline
\end{tabular}
}
\label{tab:pope}
\end{table*}

\section{Experiment}

This section will elaborate on the evaluation of our proposed Active Retrieval-Augmented large vision-language model (ARA) across different LVLMs.

\subsection{Evaluation Metrics}

\textbf{POPE}~\cite{li2023evaluating}, the Polling-based Object Probing Evaluation proposed a simplified method for evaluating object hallucinations. In this evaluation framework, the LVLM is tasked with determining the presence of specific objects in a given image, with a balanced distribution of queries for object presence and absence. The evaluation protocol encompasses three distinct sampling settings: random, popular, and adversarial, each employing unique strategies for constructing negative samples. In the random setting, objects not present in the image are chosen randomly. The popular setting selects missing objects from a high-frequency pool, whereas the adversarial setting prioritizes coexisting objects that could be easily mistaken for existing objects in the image. 
The POPE benchmark aggregates data from three diverse sources: MSCOCO~\cite{lin2014microsoft}, A-OKVQA~\cite{schwenk2022okvqa}, and GQA~\cite{hudson2019gqa}. Within each dataset, every sampling setting comprises 500 images, with 6 questions posed for each image, resulting in a total of 27,000 question-answer pairs derived from the development sets of these datasets. The evaluation predominantly assesses metrics such as accuracy, precision, recall, and F1 score.

\noindent\textbf{MME} ~\cite{yin2023survey} is a comprehensive benchmark designed specifically for evaluating LVLMs on multiple dimensions. It includes ten sub-tasks related to perception and four sub-tasks emphasizing cognition. According to previous studies~\cite{yin2023woodpecker,leng2023mitigating,chen2024halc}, we utilize the existence and count subsets for object-level hallucination evaluation, and the position and color subsets for attribute-level hallucination evaluation. Performance is quantified by accuracy and accuracy+, which is also the official evaluation method.

\noindent\textbf{MMStar} ~\cite{chen2024we}, foundational to vision-reliant multi-modal evaluation, consists of 1,500 challenge samples meticulously chosen by human experts. It is tailored to systematically benchmark six key capabilities alongside eighteen granular facets, aiming to rigorously assess the multi-modal proficiency of LVLMs using a selection of samples that have been carefully refined and equilibrated.

\noindent\textbf{MMbench} ~\cite{liu2023mmbench} is collected from multiple sources, including public datasets and the Internet, and currently, contains 2974 multiple-choice questions, covering 20 ability dimensions. Similar to MME, we focus on hallucination-related ability. Thus, we select 4 ability subsets including object localization, attribute recognition, spatial relationship, and action recognition as our evaluation benchmarks. These subsets form a comprehensive assessment of hallucinations, addressing issues at the object-level, attribute-level, and relation-level comprehensively.

\subsection{Implementation Details} 

In the case of coarse-grained retrieval, we use the COCO dataset \cite{lin2014microsoft} as the retrieval database. For fine-grained retrieval, we use the VisualGenome \cite{krishna2017visual} to build the fine-grained retrieval database as it provides a detailed caption for each grounding box in the image. CLIP-Large is used to extract image embeddings for retrieval.   
To decompose target objects from the input query, we extract key entities from the input query utilizing the large language model LLaMA2-7B \cite{touvron2023llama}. 
Grounding Dino \cite{liu2023grounding} is deployed to identify the objects in the image that correspond to the extracted entities. If the objects can not be located in the image, we only use coarse-grained retrieval for the following reasoning process.
In this paper, we implement our ARA model on three LVLMs, including LLaVA 1.5, Qwen-VL, and mPLUG-Owl2 to evaluate the effectiveness of our method. For LLaVA 1.5 and mPLUG-Owl2, we use the retrieved texts to augment, while we apply the retrieved image and text pairs to boost Qwen-VL. For these three LVLMs, we implement augmentation using 3, 4, and 5 retrieval components respectively. For active retrieval, we use the query-aware active retrieval. For the hyperparameter $\alpha$ presented in Equation 6, we apply a value of 0.8 for the POPE, MMstar, and the object-level subsets within the MME benchmark, as well as for OL, SR, and ACR in the MMbench. A value of 0.4 is utilized for the remaining evaluation subsets. For active triggering retrieval, we use query-aware active retrieval for its simplicity and effectiveness. The threshold of active triggering is determined by grid search. 

In our experiments, to have a strict comparison of our RAR and vanilla LVLMs, we consistently utilize greedy decoding instead of other probability sampling methods. With this strategy, the token with the highest probability from the post-softmax distribution is directly selected as the next token. 

\subsection{Experiment Results}

\textbf{Results on POPE.} Experimental results on POPE under the random, popular, and adversarial settings are summarized in Table \ref{tab:pope}. This benchmark mainly focuses on the object-level hallucination. 
Specifically, the performances of our ARA consistently surpass the baseline results on all of the LVLMs. 
This suggests its pivotal role in augmenting LVLMs with retrieval, thereby reducing instances of object-level hallucination. 
In a more detailed model-specific analysis, RAR demonstrates varied effects across different LVLMs. For all three LVLMs, the F1 score elevation is predominantly driven by a recall boost (e.g., up to 10 points for mPLUG-Owl2), showcasing retrieving to external knowledge base can effectively help detect object presences.

\begin{table*}[t!]
\centering
\caption{Results on the hallucination subset of MME. 
}
\resizebox{0.825\linewidth}{!}{%
\begin{tabular}{@{}lllllll@{}}
\toprule
\multirow{2}{*}{Model}        & \multirow{2}{*}{Decoding} & \multicolumn{2}{c}{\textbf{Object-level}}                                   & \multicolumn{2}{c}{\textbf{Attribute-level}}                               & \multicolumn{1}{c}{\multirow{2}{*}{Total Scores$\uparrow$}} \\
                              &                           & \multicolumn{1}{c}{\textit{Existence}$\uparrow$} & \multicolumn{1}{c}{\textit{Count}$\uparrow$} & \multicolumn{1}{c}{\textit{Position}$\uparrow$} & \multicolumn{1}{c}{\textit{Color}$\uparrow$} & \multicolumn{1}{c}{}                       \\ \midrule
\multirow{2}{*}{LLaVA-1.5} & Regular & 195.00 & 126.67 & 106.67 & 140.00 & 605.00 \\
 & RAR & \textbf{195.00} & \textbf{145.00} & \textbf{133.33} & \textbf{175.00} & \textbf{648.33} \\ \midrule
\multirow{2}{*}{Qwen-VL} & Regular & 165.00 & 145.00 & 103.33 & 185.00 & 598.33 \\
 & RAR & \textbf{170.00} & 140.00 & \textbf{118.33} & \textbf{195.00} & \textbf{623.33} \\ \midrule
\multirow{2}{*}{mPLUG-Owl2} & Regular & 170.00 & 145.00 & 76.67 & 163.33 & 555.00 \\
 & RAR & \textbf{175.00} & \textbf{148.33} & \textbf{78.33} & \textbf{163.33} & \textbf{565.00} \\ \bottomrule
\end{tabular}
}

\label{tab:mme}
%\vspace{-0.2cm}
\end{table*}

\begin{table*}[t!]
\centering
\caption{Results on the MMStar benchmark. CP (coarse perception), FP (fine-grained perception), IR (instance reasoning), LR (logical reasoning), MA (mathematics), ST (science \& technology).}
\resizebox{0.825\linewidth}{!}{%
\begin{tabular}{@{}llccccccc@{}}
\toprule
Model & Decoding & CP & FGP & IR & LR & MA & ST & \emph{Ave.} \\ \midrule
\multirow{2}{*}{LLaVA-1.5} & Regular & 0.556 & 0.264 & 0.388 & 0.276 & 0.236 & 0.208 & 0.321 \\
& RAR & \textbf{0.660} & \textbf{0.416} & \textbf{0.464} & \textbf{0.340} & \textbf{0.320} & \textbf{0.252} & \textbf{0.409} \\ \midrule
\multirow{2}{*}{Qwen-VL} & Regular & 0.540 & 0.348 & 0.460 & 0.288 & \textbf{0.292} & 0.204 & 0.355 \\
& RAR & \textbf{0.584} & \textbf{0.428} & \textbf{0.524} & \textbf{0.324} & 0.260 & \textbf{0.280} & \textbf{0.400} \\ \midrule
\multirow{2}{*}{mPLUG-Owl2} & Regular & 0.544 & 0.268 & 0.420 & 0.328 & 0.236 & 0.172 & 0.328 \\
& RAR & \textbf{0.636} & \textbf{0.396} & \textbf{0.480} & \textbf{0.392} & \textbf{0.312} & \textbf{0.252} & \textbf{0.411} \\ \bottomrule
\end{tabular}
}
\label{tab:mmstar}
\end{table*}

\noindent\textbf{Results on MME Hallucination Subset.} 
The MME subset evaluations extend beyond POPE’s scope, encompassing both object-level and attribute-level hallucination. Results in Table \ref{tab:mme} show that our ARA leads to a uniform enhancement in addressing object-level hallucination for all models, except the count score of Qwen-VL. Meanwhile, our ARA demonstrates significant improvements in attribute-level Color scores, contributing to substantial overall performance gains. These improvements are attributable to our designed coarse-to-fine retrieval paradigms, which effectively focus on the target object.

\noindent\textbf{Results on MMStar.} In a more difficult benchmark MMStar which rigorously assesses the multi-modal proficiency, as depicted in Table \ref{tab:mmstar}, our method ARA secures a notable enhancement across all six subsets. For instance, LLaVA-1.5 realizes an average improvement of 8.8 points. Unlike POPE, which may include statistical bias, answering questions in MMStar necessitates reliance on image information. Hence, our retrieval augmentation method introducing external image and text knowledge significantly bolsters model performance.

\noindent\textbf{Results on MMBench Hallucination Subset.} 
This benchmark extends our hallucination evaluation to incorporate the relation level in addition to the above object and attribute levels, providing a more holistic assessment for reducing hallucinations. As evidenced in Table \ref{tab:mmbench}, benefiting from coarse-grained retrieval, our ARA manifests discernible enhancements in the object and relation-level subsets, such as OL, SR, and ACR. For instance, our ARA model surpasses the vanilla LLaVA-1.5 with 5.08 percent points on object localization (OL).  
Concurrently, the integration of fine-grained retrieval into our model facilitates superior performance in attribute-level subsets, namely attribute recognition (ATR). 

\begin{table*}[t!]
\centering
\caption{Results on the hallucination subsets of MMbench benchmark. OL (object localization),
ATR (attribute recognition), SR (spatial relationship), and ACR (action recognition).}
\begin{tabular}{@{}llccccc@{}}
\toprule

Model & Decoding & OL & ATR & SR & ACR & \emph{Ave.} \\ \midrule
\multirow{2}{*}{LLaVA1.5} & Regular & 0.6032 & 0.8750 & 0.3785 & 0.9581 & 0.7147 \\
& RAR & \textbf{0.6540} & \textbf{0.8902} & \textbf{0.4350} & \textbf{0.9628} & \textbf{0.7293}\\ \midrule
\multirow{2}{*}{Qwen-VL} & Regular & 0.5397 & 0.8144 & 0.3842 & 0.9116 & 0.6684 \\
& RAR & \textbf{0.6413} & \textbf{0.8523} & \textbf{0.4407} & \textbf{0.9349} & \textbf{0.7271} \\ \midrule
\multirow{2}{*}{mPLUG-Owl2} & Regular & {0.5206} & {0.7955} & {0.3672} & {0.9209} & \textbf{0.6560} \\
& RAR & \textbf{0.5873} & \textbf{0.8182} & \textbf{0.4237} & \textbf{0.9581} & \textbf{0.7024} \\ \bottomrule
\end{tabular}
\label{tab:mmbench}
\end{table*}

\begin{table}[t!]
    \centering
    \caption{Ablation studies on different retrieving methods. The results are obtained on POPE MSCOCO Popular by LLaVA 1.5. For this study, we only conduct coarse-grained retrieval.}
    {
    \begin{tabular}{lcccc}
    \hline \hline
    \multirow{1}*{Methods} & Accuracy & Precision & Recall & F1 Score \\ 
    \hline
    T $\rightarrow$ T & 86.67 & 93.72 & 78.60 & 85.50 \\
    T $\rightarrow$ I & 51.60 & 50.81 & 99.87 & 67.36 \\
    I $\rightarrow$ I & \textbf{86.93} & 91.41 & 81.53 & \textbf{86.19} \\
    I $\rightarrow$ T & 86.67 & 90.92 & 81.47 & 85.94 \\ \hline \hline

    DINOv2-Base  & 86.07 & 90.37 & 80.73 & 85.28 \\
    CLIP-Base & 86.33 & 90.79 & 80.87 & 85.54 \\
    CLIP-Large & \textbf{86.93} & 91.41 & 81.53 & \textbf{86.19} \\ 

    \hline \hline
    
    \end{tabular}}
    % \vspace{-10pt}
    \label{tab:other_ablation}
\end{table}

\section{Ablation Study}

In this section, we undertake a series of ablation studies with the aim of identifying the critical elements integral to constructing a resilient and efficient retrieval-augmented large vision-language model.

\subsection{Ablation of Coarse-to-Fine Retrieval}

In this section, in order to explore the effects of different factors on coarse-to-fine retrieval in a simple and effective way, we did not use the rerank method or active trigger strategy.

\noindent\textbf{Which kind of retrieval is better?} 
Due to the multimodal characteristics of the input, there are various retrieval methods at our disposal. As shown in Table \ref{tab:other_ablation}, we attempt to use the input query and image to search for images and descriptions in the database. For this study, we only conduct coarse-grained retrieval and we use the COCO dataset as the database which contains a caption for each image. 
As a result, there are four retrieval methods available to us. 
As reflected in the results within this table, we adopt the input image as our primary retrieval modality for images in the database throughout our series of experiments.
Surprisingly, using the input query to search the database for image leads to substantially poorer outcomes, achieving only 51.6\% accuracy. 
This lower performance may stem from the fact that, despite the presence of relevant keywords, the images retrieved this way tend to be surrounded by a significant degree of noise.

Additionally, we investigate the influence of different embedding models on the efficacy of image-to-image retrieval. 
In our comparison, we examine three models: CLIP-Base/Large~\cite{radford2021learning} and DINOV2-Base~\cite{oquab2023dinov2}. Among them, we notice that CLIP-base performs better than DINO (86.33 vs 86.07) and this may be related to their different training methods. Moreover, when comparing CLIP models of various sizes, we observe that the larger-scale CLIP models contribute to more pronounced improvements in performance. Therefore, in the context of this study, CLIP-Large is selected as the image embedding model for retrieval purposes.

\begin{table}[t!]
    \centering
    \caption{Performance on the LLaVA-1.5. In this paper, we compare different kinds of retrieval schemes and regular decoding. ``I" indicates augmentation with only retrieved texts and ``I+T" means enhancing with both texts and images. 
    The experiments are performed on the popular part of POPE MSCOCO dataset. Coarse and Fine indicate coarse-grained and fine-grained retrieval, separately. Coarse+Fine means the instance-level fusion, while Coarse \& Fine corresponds to the probabilistic-level fusion.}
    {
    \begin{tabular}{ccccc}
    \hline \hline
    
    Method & Accuracy & Precision & Recall & F1  \\ 
    \hline
    Regular  & 85.53 & 95.02 & 75.0 & 83.83  \\
    Coarse (T) & 86.43 & 93.03 & 80.93 & 86.56 \\
    Coarse (I+T) & 86.93 & 91.41 & 81.53 & 86.19 \\
    Fine (T) & 86.20 & 89.46 & 82.07 & 85.61 \\
    Fine (I+T) & 86.53 & 93.08 & 78.93 & 85.43 \\
    Coarse+Fine (T) & 86.27 & 89.59 & 82.07 & 85.66  \\
    Coarse+Fine (I+T) & 86.77 & 91.88 & 80.67 & 85.91 \\
    \hline
    {Coarse \& Fine (T)} & \textbf{87.47} & 92.07 & 82.00 & \textbf{86.74} \\ 
    
    {Coarse \& Fine (I+T)} & 87.23 & 92.47 & 81.07 & 86.39 \\ 
    
    \hline \hline
    \end{tabular}}
    \label{tab:compare}
\end{table}

\begin{figure}[t!]
    \centering
    \includegraphics[width=0.8\linewidth]{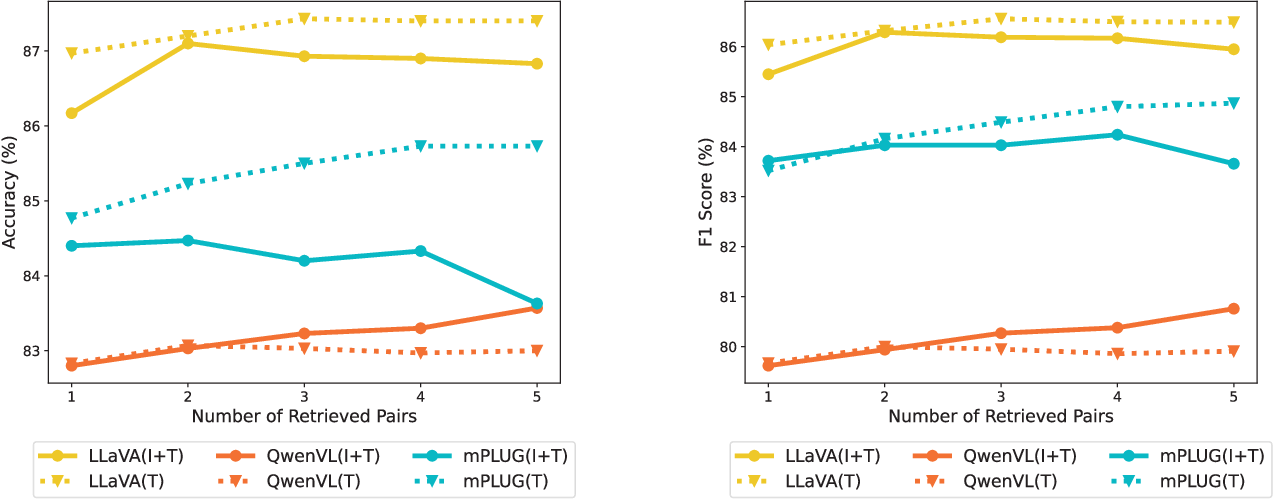}
    \caption{Performance of three LVLMs corresponding to different numbers of retrieval pairs. The performance trends of accuracy and F1 score are the same. The
    experiments are performed on the popular part of the POPE MSCOCO dataset.}
    \label{fig:num-of-pairs}
\end{figure}

\noindent\textbf{How to augment the LVLM?} After retrieving, we can obtain retrieved images and paired captions. In this section, we explore how to augment the LVLM with this external information. For a fair comparison, in this study, we fix the number of retrieved pairs to 3 for both coarse-grained and fine-grained retrieval. As demonstrated in Table 6, when using coarse-grained retrieval, we observe that the image and caption pair perform worse than the single caption (86.93 vs 87.43). 
However, the opposite results are observed in fine-grained retrieval and the Fine (I+T) achieves 86.53 compared with 86.20 of Fine (T). These results imply that
retrieval augmentation is sensitive to specific configurations. In this experiment, we do not carefully compare with Coarse (I) or Fine (I) as we have found that the LVLMs we use in our experiments are still quite weak in multi-image reasoning.

In Section 3.5, we propose a joint coarse-grained and fine-grained decoding, designed to optimally utilize the two retrieval mechanisms. 
In Table \ref{tab:compare}, we also investigate the effectiveness of probabilistic-level fusion denoted as Coarse \& Fine and instance-level fusion labeled as Coarse+Fine. 
As shown from the results, we observe that Coarse+Fine even performs than Coarse retrieval, signifying that simply integrating the coarse-grained and fine-grained retrieved results in a prompt fails to enhance performance, given the distinct granularity of the information retrieved through these two approaches. Instead, the probabilistic-level fusion Coarse \& Fine achieves better accuracy than both two kinds of retrieval. 
Thus, in our paper, we use probabilistic-level fusion for the following experiments. It is worth noting that Coarse \& Fine (T) performs better than Coarse \& Fine (I+T) for LLaVA-1.5 but Coarse \& Fine (I+T) achieves better for Qwen-VL (we will describe in Figure 3 later). This result further validates the necessity of carefully selecting retrieval configurations to obtain the most favorable experimental results.

\noindent\textbf{How much external information do we need?} 
With the above retrieval, we can obtain multiple image-caption pairs from the database. This research delves into the efficacy of varying retrieved pair configurations. 
Considering that different LVLMs may be sensitive to retrieval configurations, our study incorporates both (T) and (I+T) configurations for a thorough evaluation. The analysis is constrained to examining between one to five retrieved pairs due to the input token limitations inherent in contemporary LVLMs. 
As depicted in Figure \ref{fig:num-of-pairs}, LLaVA (T) reaches peak accuracy with three retrieved texts, whereas mPLUG (T) begins to saturate when there are four retrieval texts.
Conversely, QwenVL (I+T) demonstrates improved outcomes with a greater number of retrieval pairs; however, it still lags considerably behind the performance achieved by the other two LVLM models.

\begin{figure}[t!]
    \centering
    \includegraphics[width=0.95\linewidth]{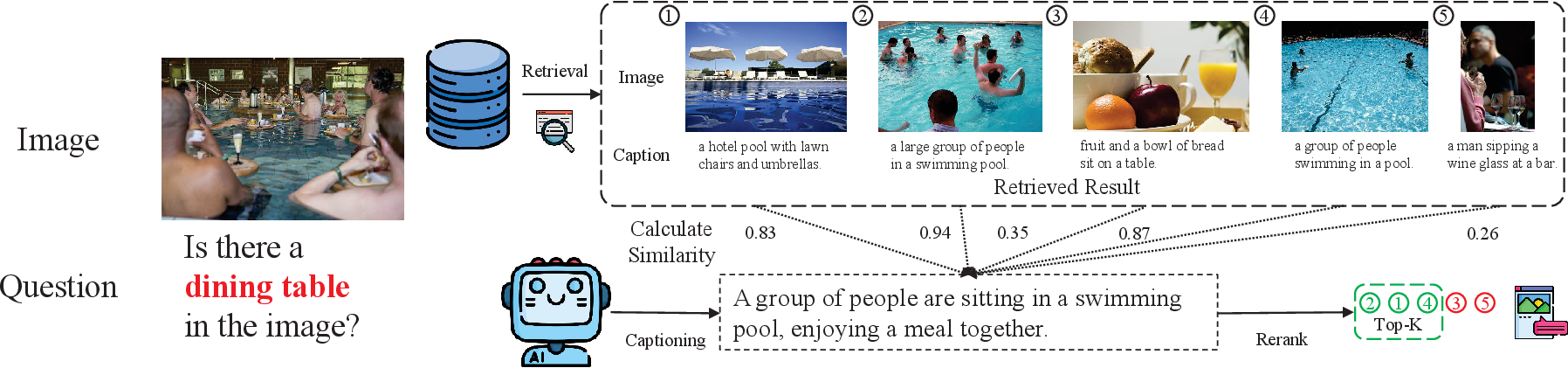}
    \caption{The detailed reranking process in our method. With reranking, the retrieved five pairs are sorted in a new order.
    This example is from the POPE benchmark and the LVLM used here is LLaVA 1.5.}
    \label{fig:rerank}
\end{figure}

\begin{table}[t!]
    \centering
    \caption{Ablation studies on different reranking methods. The results are obtained on POPE MSCOCO Popular by LLaVA-1.5. In this experiment, we incorporate a reranking strategy on top of the coarse-grained retrieval. Thus, the result of the first line (None) aligns with the results in Table 5.}
    {
    \begin{tabular}{lcccc}
    \hline \hline
    \multirow{1}*{Methods} & Accuracy & Precision & Recall & F1 Score \\ 
    \hline
    None & 86.93 & 91.41 & 81.53 & 86.19 \\

    $k$-Reciprocal \cite{zhong2017re} & 87.03 & 91.36 & 81.8 & 86.32 \\
    VSR \cite{Liu2022VisualSR} & 86.97 & 91.29 & 81.73 & 86.25 \\ \hline 

    RAR & \textbf{87.17} & \textbf{91.57} & \textbf{81.87} & \textbf{86.45} \\

    \hline
    
    \end{tabular}}
    % \vspace{-10pt}
    \label{tab:rerank}
\end{table}

\subsection{The effectiveness of Reranking Method}

To substantiate the efficacy and indispensability of reranking, we begin by providing a case study within the POPE benchmark. 
As depicted in Figure \ref{fig:rerank}, in instances where an image portrays people in a pool, leveraging CLIP for image retrieval may inadvertently introduce irrelevant content due to its predominant emphasis on visual congruence.
For instance, it might retrieve an incongruent image captioned ``fruit and a bowl of bread sit on the table." Therefore, an adept reranking methodology can proficiently filter and organize the retrieved images in a more relevant manner. In this particular scenario, our reranking strategy aligns the input image caption with those of the retrieved images. 
By assessing these similarity metrics, we are able to reorganize the pool of retrieved images into a more accurately ranked series, thereby enhancing the precision of the retrieval process.

In order to further quantify the effect of reranking and compare different reranking methods, we conduct experiments in Table \ref{tab:rerank}. 
The result reveals that the omission of re-ranking precipitates a decline in accuracy, dropping from 87.17\% to 86.93\%
This suggests that re-ranking effectively mitigates noise, particularly when employing the top three re-ordered retrieval pairs. 
For comparative purposes, we executed two alternative methodologies. The k-Reciprocal approach \cite{zhong2017re}, a prevalent offline person re-identification method, re-ranks images using the k-Reciprocal nearest neighbor algorithm. While this method offers a swifter execution by circumventing the captioning process of input images, its efficacy slightly underperforms our strategy. Additionally, we examined the employment of the external tool VSR \cite{Liu2022VisualSR}, which determines the entailment between the caption of the input image and the retrieved images. As shown in Table 7, this method gains only a slight increase in accuracy (86.93 vs 86.97). This may be due to the visual entailment model not being sufficiently versatile.

\begin{figure}[t]
    \centering
    \includegraphics[width=0.9\linewidth]{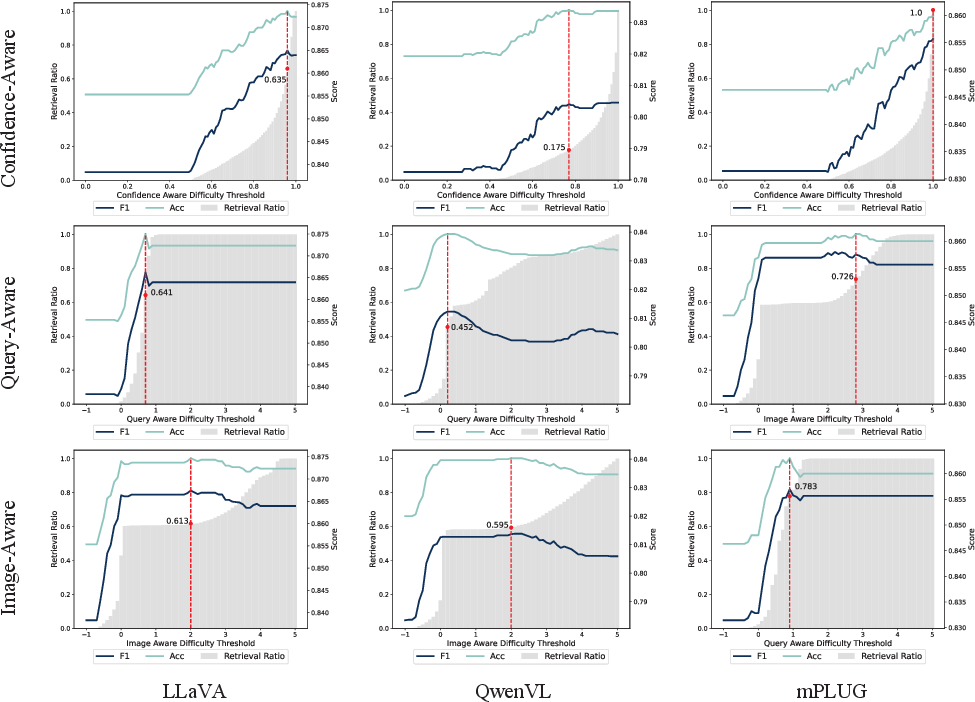}
    \caption{Different strategies for triggering retrieval on the popular part of POPE MSCOCO. The horizontal axis represents the difficulty threshold, the left vertical axis represents the proportion of queries that trigger retrieval out of all queries, and the right vertical axis shows the values of F1 score and accuracy. From top to bottom, the three lines respectively represent: confidence-aware active retrieval, query-aware active retrieval, and image-aware active retrieval.}
    \label{fig:strategies}
\end{figure}

\begin{table}[t!]
\centering
\caption{The performance comparison between our method RAR and VCD in POPE MSCOCO benchmark. VCD is the state-of-the-art method published in CVPR24. The results of VCD-sampling are from their paper. For a fair comparison with our method, we rerun VCD with greedy decoding.}
\begin{tabular}{@{}llcccc@{}}
\toprule

Setting & Decoding & Accuracy$\uparrow$ & Precision & Recall & F1 Score$\uparrow$  \\ \hline

\multirow{3}{*}{\textit{Random}} & VCD-greedy \cite{leng2023mitigating} & 85.37 & 93.24 & 76.27 & 83.90  \\

& VCD-sampling \cite{leng2023mitigating} & 87.73 & 91.42 & 83.28 & 87.16  \\
& RAR & \textbf{89.43} & \textbf{95.96} & \textbf{82.33} & \textbf{88.63} \\ \midrule

\multirow{3}{*}{\textit{Popular}} & VCD-greedy \cite{leng2023mitigating} & 83.30 & 88.75 & 76.27 & 82.04 \\

& VCD-sampling \cite{leng2023mitigating} & \text{85.38} & \text{86.92} & \text{83.28} & \text{85.06} \\

& RAR & \textbf{87.47} & \textbf{92.07} & \textbf{82.00} & \textbf{86.74} \\ \midrule

\multirow{3}{*}{\textit{Adversarial}} & VCD-greedy \cite{leng2023mitigating} & 80.00 & 82.37 & 76.33 & 79.24 \\

& VCD-sampling \cite{leng2023mitigating} & \text{80.88} & \text{79.45} & \text{83.29} & \text{81.33} \\

& RAR & \textbf{84.58} & \textbf{87.17} & \textbf{80.73} & \textbf{83.92} \\ \bottomrule
\end{tabular}
\label{tab:vcd}
\end{table}

\subsection{Ablation of Active Triggering Retrieval}

As excessive retrieval activations may lead to undue time expenditure and unnecessary retrieval, in this paper, retrieval is activated using a difficulty metric. In this ablation study, we compare three kinds of active retrieval methods, including confidence-aware active retrieval, query-aware active retrieval, and image-aware active retrieval. As shown in Figure \ref{fig:strategies}, we have drawn the proportion of queries that need to be retrieved to achieve peak performance. Firstly, it is quite intuitive, the fluctuations of confidence-aware active retrieval among different models are very significant. For mPLUG-Owl2, we even need to retrieve all the queries, which indicates that confidence-aware active retrieval is not a good indicator of whether to perform a retrieval. For both query-aware active retrieval and image-aware active retrieval, the LVLMs can better determine a trigger threshold. 
Considering that query-aware active retrieval is more concise and the process of adding noise is omitted, we use query-aware active retrieval in all our experiments. 
It is worth noting that the very slight decline in the model's performance after continuing to use retrieval beyond the trigger threshold is due to the fact that the model itself has a high level of confidence, but the retrieval has introduced some redundant noise information, which we cannot filter out through reranking.

\subsection{Comparison with State-of-the-Art Method}

In this paper, we augment LVLMs with external knowledge to mitigate hallucinations. 
To further demonstrate the effectiveness of our method, we compare it with VCD \cite{leng2023mitigating} and the results are demonstrated in Table \ref{tab:vcd}. VCD is a state-of-the-art training-free method that explores intra-model knowledge with visual contrastive decoding. From the results, we can notice that our method significantly surpasses VCD in all three settings. For instance, in popular settings, our method gains a 4.06 percent improvement. It is worth noting that VCD performs comparably with the vanilla LLaVA-1.5 when using greedy decoding. {This phenomenon is also observed in \cite{chen2024halc}}. Instead, our method significantly surpasses the vanilla LLaVA-1.5 model as presented in Tabe 1. This result further verifies the robust improvement of introducing external information when applying our ARA method.

\begin{table*}[t!]
\centering
\caption{Text generation quality comparison. Results on the MS-COCO dataset.}
% \resizebox{0.825\linewidth}{!}{%
\begin{tabular}{@{}llcccccc@{}}
\toprule
Model & Decoding & B@4  & METEOR & ROUGE-L & CIDEr & SPICE & \emph{Ave.} \\ \midrule
\multirow{2}{*}{LLaVA-1.5} & Regular                   & 22.3 & 28.0   & 50.9    & 75.3  & 22.1  & 39.7                  \\
                           & RAR                       & 
                           \textbf{25.8} & \textbf{28.3}   & \textbf{53.2}    & \textbf{94.2}  & \textbf{22.2}  & \textbf{44.7}   
                           
                           \\ \midrule
\multirow{2}{*}{Qwen-VL}   & Regular                   & 25.3 & 25.5   & 50.3    & 96.4  & 19.8  & 43.4                  \\
                           & RAR                       & \textbf{40.8} & \textbf{29.8}   & \textbf{60.5}    & \textbf{137.2} & \textbf{23.4}  & \textbf{58.3}     
                           
                           \\ \midrule
\multirow{2}{*}{mPLUG-Owl2}   & Regular                   & 28.4 & 27.3   & 54.1    & 96.5  & 19.0  & 45.1                  \\
                           & RAR                       & \textbf{32.1} & \textbf{28.5}   & \textbf{56.3}    & \textbf{103.7} & \textbf{20.8}  & \textbf{48.3}                  \\ 
\bottomrule
\end{tabular}
% }
\end{table*}

\subsection{Text Generation Quality Evaluation}

In this section, {we evaluated the effectiveness of our method in improving text generation quality on the MS-COCO validation set\footnote{We use the official COCO evaluation package.}. As shown in Table 9, all models LLaVA, Qwen-VL, and mPLUG exhibited significant improvements after applying our method, with an average improvement of 7.7\% across the three models. Notably, the improvement was most pronounced for Qwen-VL, with an average increase of approximately 14.9\%.}

\subsection{Qualitative Results}

\begin{figure}[t!]
    \centering
    \includegraphics[width=0.95\linewidth]{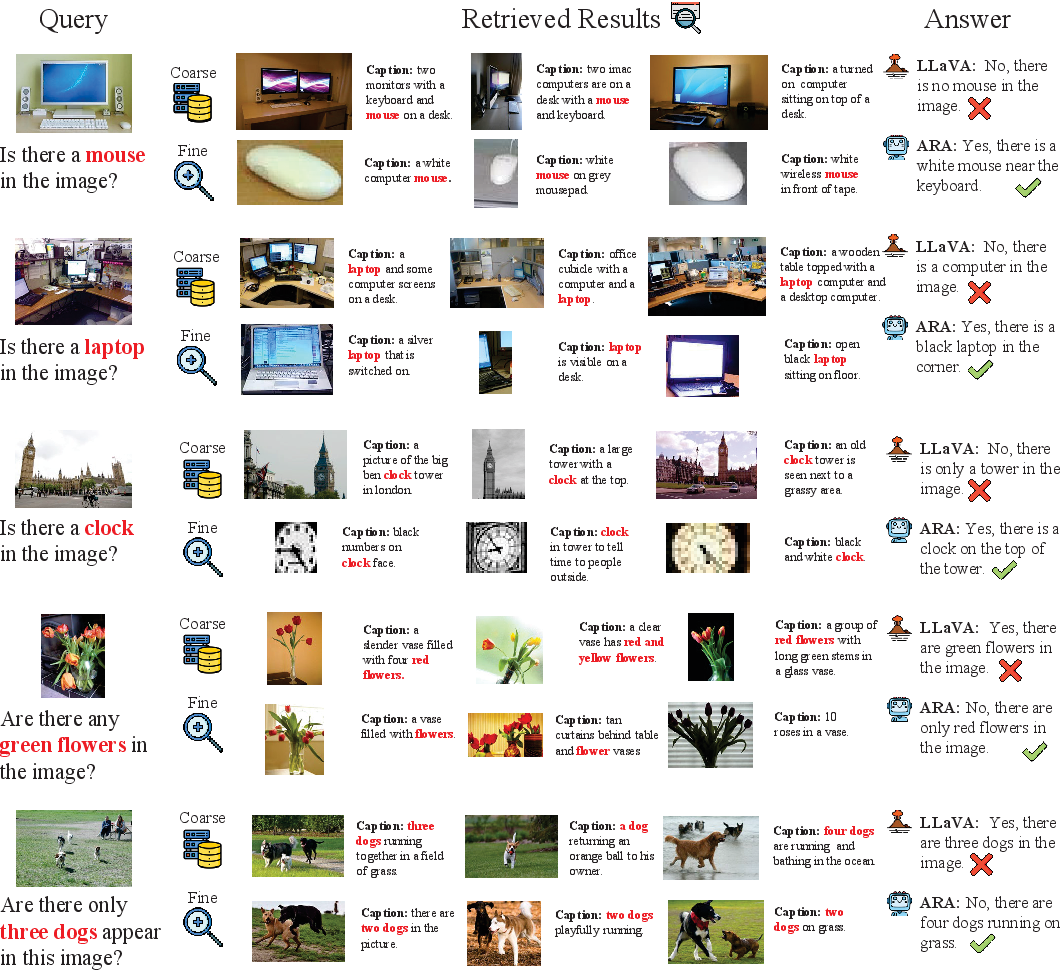}
    \caption{Examples from POPE benchmark and MME. We present both coarse-grained and fine-grained retrieval for each instance.}
    \label{fig:case}
\end{figure}

To qualitatively verify the effectiveness of our ARA method on downstream tasks, we presented five examples from the POPE MSCOCO dataset and MME benchmark. As shown in Figure \ref{fig:case}, we present three image-caption pairs for both coarse-grained retrieval and fine-grained retrieval. In the above three examples of determining the existence of objects, fine-grained retrieval helps to increase the model's existence information about specific objects. For the last two examples, the coarse-grained retrieval contributed effectively to capturing the overall information of the images. Thus, with both kinds of retrieval, our RAR model can effectively increase the accuracy compared with vanilla LLaVA-1.5.

\section{Conclusion}

In this paper, we propose a novel framework to augment LVLMs with external knowledge by introducing an
Active Retrieval-Augmented large vision-language model (ARA) for mitigating hallucinations. Specifically, our proposed ARA is grounded in three critical dimensions for mitigating hallucinations in LVLMs, including coarse-to-fine retrieval for dissecting the retrieval targets, pinpointing the most effective retrieval methods, and triggering the retrieval process properly. 
Our findings indicate that by utilizing fitting retrieval mechanisms and timing the retrieval judiciously, we can effectively mitigate the hallucination problem. Our extensive experiments across four benchmarks and three widely-used LVLM confirm ARA’s efficacy in reducing hallucination.

\bibliographystyle{ACM-Reference-Format}
\bibliography{tomm-rag}

\end{document}